\documentclass[runningheads]{llncs}

\usepackage{eccv}

\usepackage{eccvabbrv}

\usepackage{graphicx}
\usepackage{booktabs}
\usepackage{array}
\usepackage{multirow}
\usepackage[accsupp]{axessibility}  

\usepackage[pagebackref,breaklinks,colorlinks,citecolor=eccvblue]{hyperref}

\usepackage{orcidlink}

\usepackage{xspace}
\usepackage{bm}
\usepackage{subcaption}
\usepackage{ulem}
\usepackage{graphicx}

\newcommand{\Figref}[1]{Fig.\ref{#1}}

\newcommand{\sgil}{\textit{Spatial Guidance Injector}}
\newcommand{\sgis}{\textit{SGI}}

\newcommand{\dcls}{\textit{DCL}}

\renewcommand{\baselinestretch}{0.99}
\begin{document}

\title{ECNet: Effective Controllable Text-to-Image Diffusion Models}

\titlerunning{ECNet}

\author{Sicheng Li\inst{1} \and
Keqiang Sun\inst{2} \and
Zhixin Lai\inst{3} \and
Xiaoshi Wu\inst{2} \and
Feng Qiu\inst{4} \and 
Haoran Xie\inst{1} \and
Kazunori Miyata\inst{1} \and
Hongsheng Li\inst{2}
}
\authorrunning{Li et al.}

\institute{Japan Advanced Institute of Science and Technology \and
The Chinese University of Hong Kong \and
Cornell University \and
Shanghai Jiao Tong University\\
}

\maketitle

\begin{abstract}

The conditional text-to-image diffusion models have garnered significant attention in recent years. However, the precision of these models is often compromised mainly for two reasons, ambiguous condition input and inadequate condition guidance over single denoising loss.
To address the challenges, we introduce two innovative solutions. Firstly, we propose a Spatial Guidance Injector (SGI) which enhances conditional detail by encoding text inputs with precise annotation information. This method directly tackles the issue of ambiguous control inputs by providing clear, annotated guidance to the model.
Secondly, to overcome the issue of limited conditional supervision, we introduce Diffusion Consistency Loss (DCL), which applies supervision on the denoised latent code at any given time step. This encourages consistency between the latent code at each time step and the input signal, thereby enhancing the robustness and accuracy of the output. The combination of SGI and DCL results in our Effective Controllable Network (ECNet), which offers a more accurate controllable end-to-end text-to-image generation framework with a more precise conditioning input and stronger controllable supervision. We validate our approach through extensive experiments on generation under various conditions, such as human body skeletons, facial landmarks, and sketches of general objects. The results consistently demonstrate that our method significantly enhances the controllability and robustness of the generated images, outperforming existing state-of-the-art controllable text-to-image models.


\end{abstract}


\begin{figure}[t]
  \centering
  \includegraphics[width=1.0\textwidth]{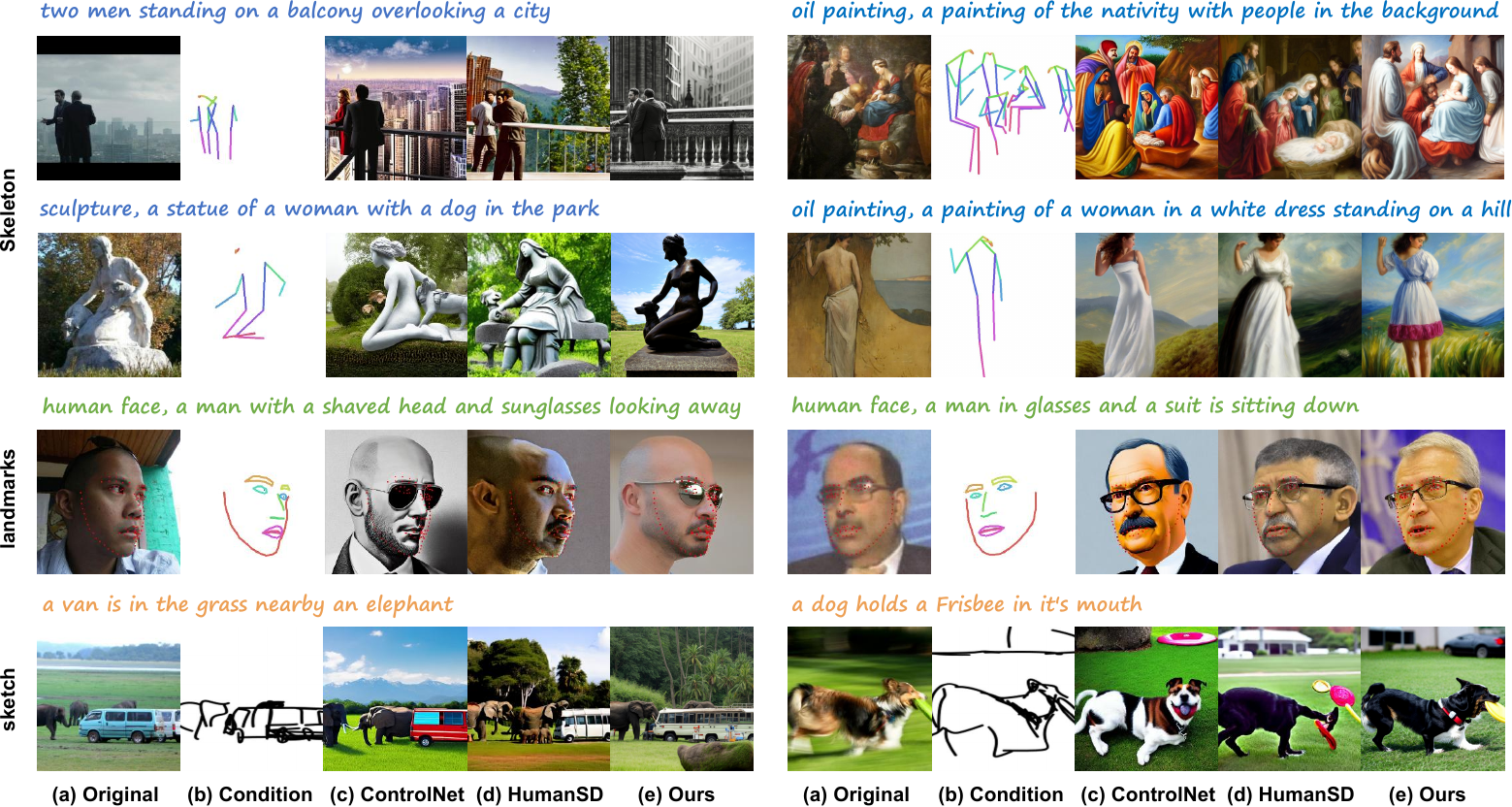}
   \caption{The core work of this paper is to design a general framework for supervised training of diffusion models, and enhancing the controllability of text-to-image diffusion models. The figures show three categories conditions, skeleton (rows I and II), facial landmark (row III), and sketch (row IV). Each category includes five sets of images, where each set displays: (a) the original image used for reference; (b) the pose (facial contour or sketch) image derived from the original image as the control condition; (c) results generated by ControlNet; (d) results generated by the baseline model, HumanSD, for comparison; (e) results generated by ECNet (our model). Compared to ControlNet and HumanSD, our model ECNet exhibits superior capabilities and robustness in image generation with control across all categories.}
  \label{teaser}
\end{figure}
\section{Introduction}
\label{sec:intro}

Controllable Image Generation represents a seminal area of inquiry within the domains of computer vision and deep learning, as evidenced by recent contributions~\cite{zhang2023adding,mou2023t2iadapter, ju2023humansd, piao2021inverting, suncontrollable, sun2022cgof++}. The capacity to synthesize images that conform to predetermined conditions not only extends the frontiers of conventional image synthesis methodologies but also serves an array of application-specific demands. Such technological advancements are of paramount importance in disciplines such as virtual reality, film production, and fashion design, where the capability to automate the creation of images tailored to particular themes can substantially augment efficiency and mitigate expenses.  

Due to the unparalleled performance of diffusion models in text-to-image generation, they have outperformed the results generated by Generative Adversarial Networks (GANs)~\cite{gan, mustgan21_gan, nted22_gan} and Variational Autoencoders (VAEs)~\cite{vae, kpe22_vae} in image generation. The forefront of controllable diffusion models, epitomized by ControlNet~\cite{zhang2023adding} and T2I-Adapter~\cite{mou2023t2iadapter}, has realized a measure of control in the generation of images.
Both models enhance the Stable Diffusion (SD) framework with a new, trainable module, making image generation more adaptable to specific requirements. This supplementary branch incorporates diverse constraints, including skeletons and sketches, throughout the image generation process, thereby markedly augmenting the controllability of the foundational SD model. Recent advancements such as HumanSD~\cite{ju2023humansd} and Composer~\cite{huang2023composer} have adopted a refined approach that concatenates extra conditions to the noisy latent embeddings as the input of the SD U-Net module. This approach facilitates a more stable training regimen and reduces the complexity of the learning objective, consequently bolstering the model's controllability and robustness.

Despite the advances in controllability for the state-of-the-art diffusion models, there remains substantial scope for improvement. For instance, the leading skeleton-guided diffusion model, HumanSD, struggles with generating images in intricate scenes, dynamic actions, and nuanced details. This limitation primarily stems from the limited controllability inherent in the end-to-end training methodology, which includes issues such as the ambiguity of condition inputs and the lack of comprehensive conditional supervision beyond a singular denoising loss. To resolve the main issues of the state-of-the-art models, we introduce two innovative solutions.

\begin{figure}[t]
  \centering
   \includegraphics[width=1.0\linewidth]{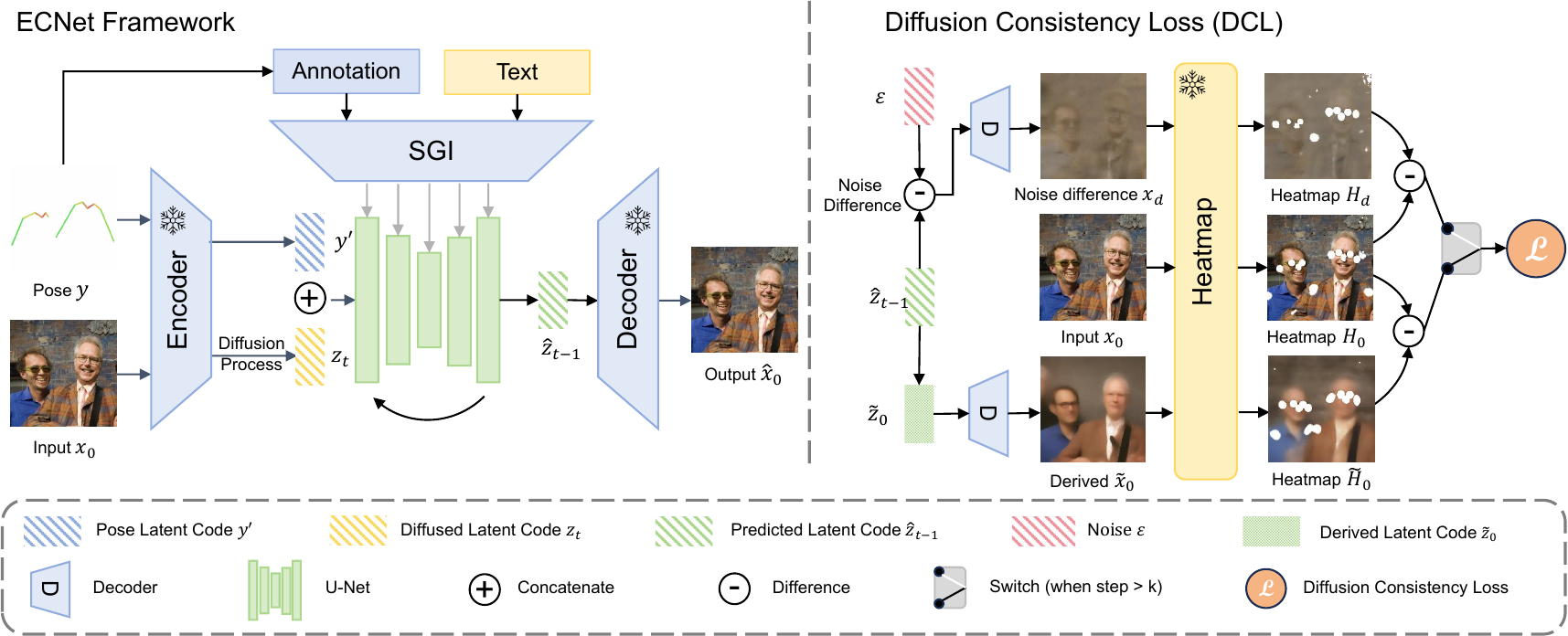}
   \caption{The framework and its loss design are illustrated using the task of skeleton control as an example. our model encodes the skeleton image into a latent code via a VAE to obtain a pose latent code. This code combines with diffusion's noise code as input for a U-Net. Additionally, Our $\sgis$ module further combines corresponding pose annotations and text, integrating them into the U-Net layers. During the training phase, we enhance the conditional generation capabilities of the diffusion model by introducing $\dcls$. $\dcls$ targets heatmap disparities between estimated and input images, using dual-stage loss to impose consistency supervision throughout the duffusion process. $z$ represents the latent code and $x$ denotes the image decoded from $z$. Please refer to~\cref{sec:method} for more details.}
   \label{framework}
\end{figure}

Firstly, we introduce the Spatial Guidance Injector (SGI) to incorporate precise annotation information as a condition, rather than solely depending on the condition image, as illustrated in \Figref{framework}. Annotations serve as an effective means to define human postures and facial orientations. Thus, the combination of image and annotation furnishes a more comprehensive and nuanced understanding of the condition. For example, while the condition image provides the global structure, the annotations offer more precise and detailed context. We propose a novel and effective approach to combine the image condition and annotation condition with text condition: the image is processed through a U-Net architecture for global condition feature extraction, whereas annotations are combined with textual conditions through the $\sgis$ architecture, enriching the context with finer details.

Secondly, we introduce Diffusion Consistency Loss (DCL) by firstly using the denoised latent code in loss calculation, rather than focusing on the noise itself in traditional diffusion models, shown in \Figref{framework}. This denoised latent code supervision helps guide the model more accurately toward the ground truth because it compares to something much closer to the output. Moreover, currently, any methods estimating images from the predicted denoised latent code struggle to maintain high fidelity across all time steps of diffusion process, as shown in \Figref{timestep}. This makes it challenging to apply stable and consistent supervision on the latent code of diffusion model. To address this, our introduced $\dcls$ adopts dual-stage loss formulations adaptable to every phase of the denoising process. The dual loss formulation enhances supervision by offering more detailed and precise guidance, and it also contributes to a more stable training process.

The main contributions of this paper are: 

\begin{itemize}

\item The introduction of ECNet, an innovative framework for controllable image generation that integrates annotation data with textual descriptions via our pioneering Spatial Guidance Injector (SGI) architecture. This method provides enhanced input control, leading to greater contextual depth and improved controllability over the generated images.

\item The development of a novel Diffusion Consistency Loss (DCL) within ECNet. $\dcls$ is the first to utilize denoised latent code for supervision and incorporates dual loss formulations tailored for different stages of the training process. This significantly boosts the model's controllability, robustness of its outputs. 

\item The efficacy and efficiency of the ECNet framework are validated through various evaluation metrics across multiple domains, including skeletons, landmarks, and sketches, shown in \Figref{teaser}. The performance of ECNet surpasses previous state-of-the-art models in a fair experimental setting.

\end{itemize}

\section{Related Work}
\label{sec:related}
\subsection{Text-to-Image Diffusion Model}

Diffusion models have emerged as the new state-of-the-art in the realm of deep generative models. Surpassing Generative Adversarial Networks(GANs)\cite{goodfellow2014generative} in image generation tasks, diffusion models have achieved cutting-edge results in image synthesis. Benefiting from the remarkable ability of large-scale language models, such as CLIP\cite{radford2021learning}, to encode textual inputs into latent vectors, diffusion models have demonstrated astonishing capabilities in text-to-image generation tasks. For instance, one of the earliest text-to-image diffusion models, Glide\cite{nichol2022glide}, is a text-guided diffusion model that also supports image super-resolution generation and editing. Imagen\cite{saharia2022photorealistic}, a text-to-image architecture, discovered significant improvements using a pretrained large-scale text-only encoder and introduced a new Efficient U-Net structure. Latent Diffusion Model(LDM)\cite{rombach2022highresolution} was the first to propose conducting diffusion and reverse diffusion in feature space, significantly enhancing efficiency, and introduced the use of cross-attention to embed conditional information, allowing for more flexible incorporation of conditions. Stable Diffusion is a large-scale implementation of latent diffusion, designed for text-to-image generation. However, all these models typically only take text as input, making it challenging for precise image control, such as target positioning and posture control. Consequently, subsequent works have seen the emergence of numerous studies focusing on controlled image generation using diffusion models.

\subsection{Controllable Diffusion Model Generation}
In this section, we primarily focus on controllable diffusion model generation, specifically on how to incorporate additional conditions into text-to-image models, such as bounding boxes, human poses, and sketches. Among the most influential works in this area are ControlNet\cite{zhang2023adding} and T2I Adapter\cite{mou2023t2iadapter}. Both of them fix the original weights of Stable Diffusion and use an additional trained branch to modify the embeddings in the U-Net for guiding generation. We refer to this approach as the dual-branch diffusion model. This method supports a variety of conditions, including human poses, Canny Edge Maps, and more, enabling flexible image generation control. 

Additionally, there has been a surge of recent works targeting different tasks. Uni-ControlNet\cite{zhao2023unicontrolnet} and Composer\cite{huang2023composer} address image generation control under multiple conditions, considering the interrelations between different conditions. They categorize conditions into local and global, adopting different methods of integration depending on the type of condition. Both LayoutDiffusion\cite{zheng2023layoutdiffusion} and GLIGEN\cite{li2023gligen} use bounding boxes (bbox) as conditions for controlled generation. LayoutDiffusion integrates encoded bbox information into the U-Net, merging image and layout features for controlled generation. In contrast, GLIGEN adds new attention layers to handle the fusion of bbox and text without altering the original weights of Stable Diffusion, endowing the model with the capability to control generation using bbox.

In the task of human pose control, HumanSD\cite{ju2023humansd} proposed an alternative approach for adding conditions. This study combined the pose image embedding with the noisy image embedding as the input to the U-Net for training, showing superior pose control capabilities compared to dual-branch diffusion models. The article also improved the original SD loss by incorporating a weight more focused on human posture, thereby guiding the model to generate results more aligned with the pose conditions. Although the loss was optimized, this method essentially still achieves control by adjusting the input method.

In the task of sketch-to-image, \cite{voynov2022sketchguided} proposed a new control scheme. This study trained a mapping network to align U-Net features as closely as possible with sketch features. By calculating the error between the mapped results of U-Net features and the input sketch, reverse gradients are obtained to guide the update of U-Net weights, making the generated results more aligned with the sketch conditions. This approach achieves controlled generation through supervision and is more logical compared to modifying the conditional input. However, due to its reliance on a pre-trained mapping network, this method lacks versatility.

\section{Preliminaries and Motivation}\label{subsec:mov}

This section discusses the issues of existing methods and the inspiration of ECNet. These methods uniformly adopt the Latent Diffusion Model (LDM) as their foundational framework, capitalizing on its high trainability and exceptional generative quality, while employing various control schemes. These methods are introduced in Section \ref{subsec:intro}. Subsequently, Section \ref{subsec:issue} elucidates the problems present in these methods and the motivation behind designing ECNet.

\subsection{Preliminary Introduction}\label{subsec:intro}

The training process of the Diffusion model is conceptualized as a standard diffusion process, where an input latent code $z_0$ incrementally acquires noise over t time steps, transitioning into a latent code close $z_t$ approximating random noise. This process is mathematically articulated as:
\begin{equation}
\begin{gathered}
    z_{t} = \sqrt{\bar\alpha_{t}} z_{0} + \sqrt{1 - \bar\alpha_{t}} \epsilon, \quad \epsilon \sim \mathcal{N}(0, I)
\end{gathered}
\label{forward process}
\end{equation}
where $\bar\alpha_{t}$ denotes a predetermined noise level coefficient, $\epsilon$ represents noise drawn from a standard normal distribution, and $t$ signifies the time step.

During the denoising phase, the model learns to predict the input latent code $z_0$ from diffused latent code $z_t$, a process achieved by optimizing the following objective function:
\begin{equation}
\begin{gathered}
    \mathcal{L} = \mathop \mathbb{E} \limits_{t, z, \epsilon}\left[ \|\epsilon - \epsilon_\theta(z_t, t)\|^2 \right],
\end{gathered}
\label{reverse process}
\end{equation}
where $\epsilon_\theta$ is the noise predicted by the model, to minimize the discrepancy between the predicted noise and the actual noise. Through this mechanism, the Stable Diffusion Model effectively learns the data distribution and generates high-quality images.

In the realm of text-to-image diffusion model, the current state-of-the-art method of human pose control, HumanSD, introduces heatmap-guided denoising loss, as Equation~\ref{eq:heatmap_guided_denoising_loss},  It integrates a heatmap weight into the diffusion model's original loss, prioritizing relevant pixels for enhanced denoising. This approach outperforms models like ControlNet in pose accuracy and image quality under complex guidance.
\begin{equation}
\begin{gathered}
    \small
    \mathcal{L}_\text{h} = \mathop \mathbb{E} \limits_{t, z, \epsilon}\left[   \left\| W_a \cdot  \left( \epsilon -\epsilon _{\theta}\left( \sqrt{\bar{\alpha}_t}z_0+\sqrt{1-\bar{\alpha}_t}\epsilon, c, t \right)\right) \right\| ^2 \right], \\
    W_a = 1 + \lambda H_{dff}, \\
\end{gathered}
\label{eq:heatmap_guided_denoising_loss}
\end{equation}
where a weight $W_{a}$ is integrated into the original diffusion model. $W_a$ acts as a conditional guide for the diffusion process, $H_{dff}$ means the heatmap weight, $\lambda$ means a constant term controlling the weight guidance strength.

Inspired by another work~\cite{voynov2022sketchguided}, a novel method controls diffusion model inference via sketches, differing from typical condition-based control. It utilizes a pre-trained edge predictor during training for mapping noisy image features to edge maps. At each denoising step $t$, features are input into a latent edge predictor to estimate edge maps, with the similarity gradient between predicted and true edges guiding denoising, as shown in Equation~\ref{eq:updateloss}. This edge guidance ensures synthesized images closely align with the target edges.

\begin{equation}
\begin{aligned}
    \hat{z}'_{t-1} &= \hat{z}_{t-1} +  k \cdot \nabla_{z_t} Loss \\
\end{aligned}
\label{eq:updateloss}
\end{equation}
where $\hat{z}_{t-1}$ represents predicted latent code at time step $t-1$, $\hat{z}'_{t-1}$ denotes the predicted latent code after guided by the gradient, $Loss$ denotes the calculated edge loss, and $k$ governs the intensity of the guidance exerted by the loss.

\subsection{Motivation}\label{subsec:issue}

Existing SD-based control models predominantly use condition images as their primary input for control. However, relying entirely on image features for control conditions does not provide adequate guidance, leading to certain details in the generated images being controlled imprecisely.
In contrast to images, annotations include sequences and coordinates, complementing the global control provided by textual conditions, giving a more detailed guide for image generation. Therefore, we suggest that incorporating image annotation information into traditional textual conditions can significantly improve the controllability of the generated outcomes.

Beyond the issue of insufficient conditional inputs, existing diffusion models also suffer from a lack of supervision on conditions or from employing ineffective supervisory methods. For instance, ControlNet uses traditional stable diffusion loss without any condition-based supervision. While HumanSD uses a heatmap-guided weighted loss to strengthen the new structure-aware condition by weighting original diffusion loss with an estimated noise difference heatmap. However it still applied to the supervision of noise without directly supervision over the latent code.
To develop a supervision on the denoised latent code at any time steps of the diffusion model, we explore a novel method to estimate image during the denoising process.

\begin{figure*}[t]
    \centering
    \begin{subfigure}[b]{0.48\textwidth}
        \includegraphics[width=\textwidth]{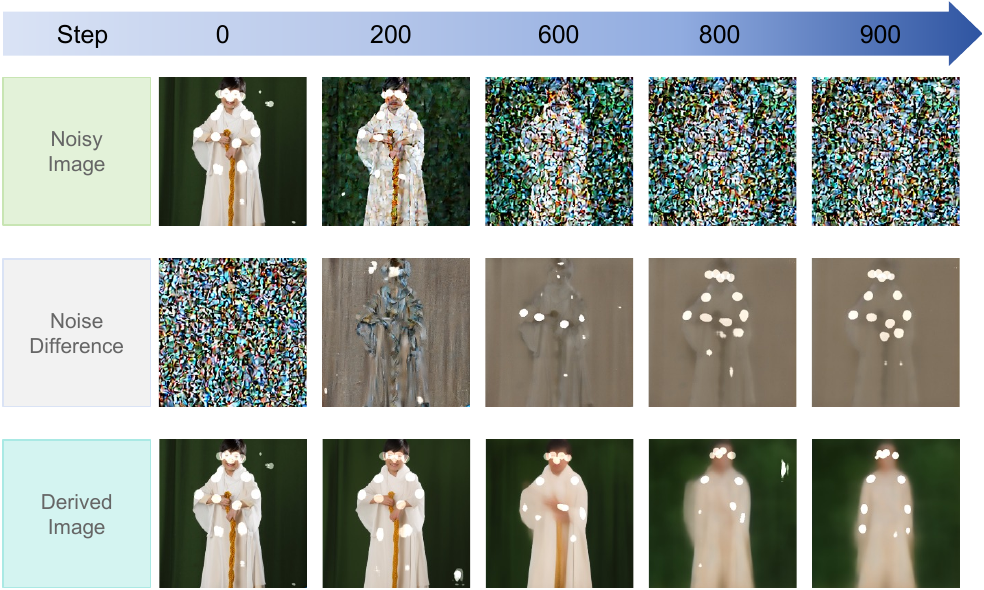}
        \caption*{}
        \label{timestep_a}
    \end{subfigure}
    \begin{subfigure}[b]{0.48\textwidth}
        \includegraphics[width=\textwidth]{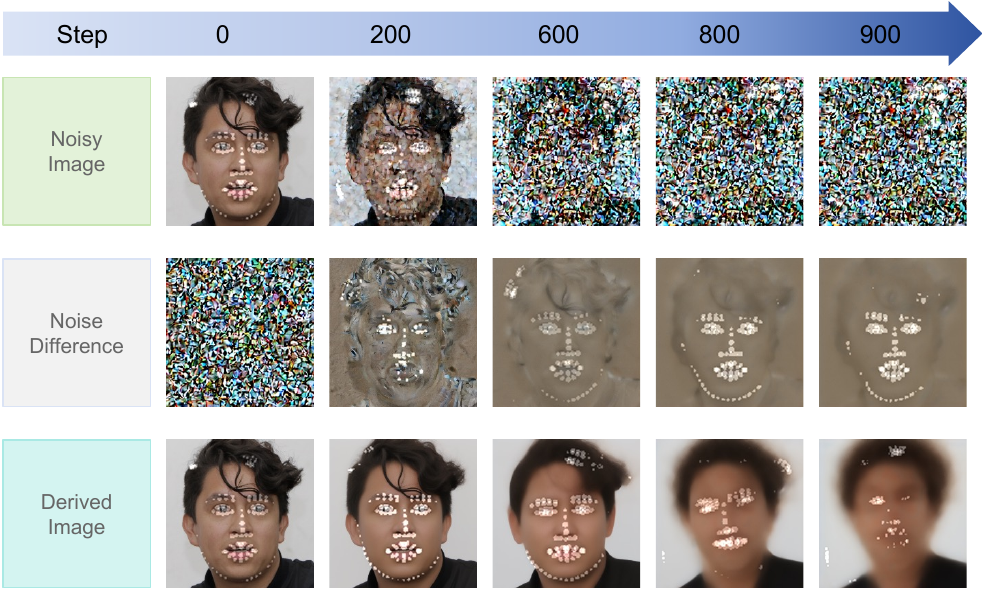}
        \caption*{}
        \label{timestep_b}
    \end{subfigure}
    \caption{The decoded images of pose and face. First row: The process of incrementally adding noise to the original image over time steps; second row: the noise difference; third row: denoised results derived from the predicted noise latent code. White areas show the detected keypoint heatmaps.}
    \label{timestep}
\end{figure*}

The diffusion process is typically fixed and employs a predefined variance schedule, allowing for the sampling of ${z}_t$ at any time step $t$ directly from $z_0$, as illustrated by Equation \ref{forward process}. Consequently, we can deduce the result as shown in Equation \ref{eq:x_0'}.
\begin{equation}
\begin{aligned}
    \Tilde{x}_0 &= \frac{\hat{x}_t - \sqrt{1 -\bar{\alpha}_t}\varepsilon}{\sqrt{\bar{\alpha}_t}}
\end{aligned}
\label{eq:x_0'}
\end{equation}
where $\hat{x}_t$ represents the model's predicted noisy image and $\Tilde{x}_0$ refers to the denoised image directly derived from the predicted noisy image at time step $t$, which is shown in the third row of \Figref{timestep}. Note that in our paper, $z$ is used to represent the latent code, and $x$ denotes the image decoded from the latent code.
It is observed that global image features, including the separation of foreground and background and the structure of the subject's pose, are primarily produced in the initial phases of the denoising process. On the other hand, the generation of local features predominantly occurs during the later stages.

Therefore, an intuitive approach involves supervision between the heatmap features of derived image $\Tilde{x}_0$ and the input image. This supervision can dynamically adjust its intensity based on time steps. Specifically, during the early time steps, the predicted image more closely resembles the original image, leading to a reduced error in heatmap detection. In contrast, at later time steps, as the fidelity of the predicted image to the original decreases, the resulting errors in heatmap prediction provide stronger supervision. This adaptable variation in supervision intensity can adapt the image generation process of the diffusion model.

However, due to the lower quality of the derived images $\Tilde{x}_0$ at larger time steps, significant errors in keypoint detection results occur, thereby diminishing the accuracy of supervision. 
To mitigate this challenge, we introduce noise difference images, as shown in the second row of \Figref{timestep}, which provides more accurate keypoint heatmaps at larger time steps. ensuring more effective and robust conditional supervision during these later stages of the denoising process.


\section{Method}
\label{sec:method}
\subsection{Diffusion Consistency Loss}\label{subsec:loss}

Inspired by existing studies, we propose the integration of additional latent code supervision into the general loss structure applicable for classifier-free guidance, to enhance the generation accuracy. The entire loss $\mathcal{L}_{e}$, as depicted in Equation~\ref{totloss}, is divided into two primary components, termed the weighted SD loss $\mathcal{L}_{h}$ and Diffusion Consistency Loss $\mathcal{L}_{DC}$. 

\begin{equation}
\begin{gathered}
    \mathcal{L}_{e} = \mathcal{L}_{h} + \alpha \mathcal{L}_{DC}
\end{gathered}
\label{totloss}
\end{equation}

According to Section~\ref{subsec:issue}, the construction of $\mathcal{L}_{DC}$ adopts distinct supervision strategies during two phases of the diffusion process as illustrated in Equation~\ref{suploss}. This loss design harnesses the high fidelity of the noise difference image and the derived image $\Tilde{x}_0$ at different timesteps, providing precise supervision for the training process.

\begin{equation}
\begin{aligned}
    \mathcal{L}_{DC} &= 
    \begin{cases} 
        \mathcal{L}_{drv} & \text{if } t < k \\
        \mathcal{L}_{dff} & \text{if } t \geq k ,
    \end{cases} \\
    \mathcal{L}_{drv} &= |H_{inp} - H_{drv}|, \\
    \mathcal{L}_{dff} &= |H_{inp} - H_{dff}|
\end{aligned}
\label{suploss}
\end{equation}

where $H$ represents the heatmap features of images decoded from latent code. $H_{inp}$, $H_{drv}$, and $H_{dff}$ means the heatmap features of the input images, the derived images $\Tilde{x}_0$, and the noise difference images. $\mathcal{L}_{drv}$ and $\mathcal{L}_{dff}$ are both L1 losses, $\mathcal{L}_{drv}$ is active for \(t < k\), emphasizing the alignment between input image and $\Tilde{x}_0$ at earlier time steps.
$\mathcal{L}_{dff}$ is relevant for \(t \geq k\), focusing on mitigating the negative impact caused by keypoint detection errors in lower-quality $\Tilde{x}_0$ at later time steps. $k$ is set to 600 based on our experience, and $\alpha$ is a constant item to control the supervision intensity.
The working principle of $\mathcal{L}_{DC}$ is the same as Equation~\ref{eq:updateloss}.

Another loss component $\mathcal{L}_{h}$ is shown in Equation~\ref{eq:heatmap_guided_denoising_loss}, We retained this weight guidance loss derived from HumanSD.

\subsection{Spatial Guidance Injector}\label{subsec:condition}

Previous pose control models based on SD employed skeletal images to incorporate pose conditions, utilizing a VAE module to process these skeletal images for positional information, ensuring alignment of pose conditions with the latent embedding of input images. However, we suppose that extracting image features to derive pose information is rather indirect. In contrast, the keypoint annotation embedded within skeletal images offers more direct spatial information for pose representation. Moreover, we observed that textual conditions typically do not encompass specific details such as the number of objects or joint positions. Given those, we propose integrating the keypoint annotations as an additional condition to the existing posture image and textual conditions. Specifically, each image is processed to extract keypoint annotations, which are then refined through a sequence of operations including padding, tokenization, masking, and embedding. Simultaneously, text embeddings are produced using the CLIP encoder. To synthesize the visual and textual information, we employ a self-attention mechanism on the annotations and integrate the results with the text embeddings via a cross-attention module. This integrated module is called \sgil(\sgis), as the Equation~\ref{eq:sgi}. The $\sgis$  facilitates a more sophisticated understanding of the multimodal annotation data. 

\begin{equation}
\begin{aligned}
&\text{SGI} = \text{softmax}\left(\frac{\mathbf{W}_Q C(t)(\mathbf{W}_K A(a))^\top}{\sqrt{d_k}}\right)(\mathbf{W}_V A(a)) + C(t) \\
\end{aligned}
\label{eq:sgi}
\end{equation}

Where, $A(a)$ symbolizes the Self-Attention mechanism applied to the annotations. $C(t)$ denotes the frozen CLIP encoder that extracts meaningful textual features from prompts.
$\mathbf{W}_Q$, $\mathbf{W}_K$, and $\mathbf{W}_V$ are the weight matrices for the Query, Key, and Value in the attention mechanism, respectively. These matrices transform the inputs into representations suitable for generating attention scores, and $d_k$ is the dimension of the key vectors.
The softmax function applied to the cross-attention result between $A(a)$ and $C(t)$, produces a distribution representing the attention weights.
The final output is derived by multiplying the result of the softmax function with the Value matrix ($\mathbf{W}_V A(a)$) and adding it to $C(t)$. This process effectively merges contextual information from both annotations and text, thereby providing a more semantically rich input to the model.

\section{Experiments}

\begin{table}[t]
\centering
\caption{Quantitative comparisons between ECNet and other SD-based models. We conduct experiments on ECNet for two primary tasks: human skeleton control and facial landmark control. The results indicate that ECNet outperforms previous SD-based models in both tasks.}
\begin{tabular}{c c c c c c| >{\centering\arraybackslash}m{30pt}>{\centering\arraybackslash}m{37pt}>{\centering\arraybackslash}m{30pt}}
\hline
\multirow{2}{*}{Model} &\multicolumn{5}{c|}{Pose Performance Metrics} & \multicolumn{3}{c}{Face Performance Metrics} \\
 & AP(\%)$\uparrow$ & CAP(\%)$\uparrow$ & PCE$\downarrow$ & CLIPSIM$\uparrow$ & FID$\downarrow$ & NME$\downarrow$ & CLIPSIM$\uparrow$ & FID$\downarrow$ \\
\hline
ControlNet~\cite{zhang2023adding} & 19.06 & 60.14 & 1.86 & 32.58 & 4.79 & 0.79 & 29.11 & 6.09 \\
HumanSD~\cite{ju2023humansd} & 33.15 & 59.38 & 1.43 & \textbf{32.63} & \textbf{4.74} & 0.971 & \textbf{29.89} & 3.95 \\
ECNet (Ours) & \textbf{43.31} & \textbf{62.76} & \textbf{1.35} & 32.28 & 4.89 & \textbf{0.657} & 29.46 & \textbf{3.21} \\
\hline
\end{tabular}
\label{tab:combined_metrics}
\end{table}

In this chapter, we evaluate the performance of the ECNet framework across various tasks, including skeleton control, facial landmarks control, and sketch control. In Section~\ref{subsec:compare}, the results indicate that ECNet surpasses the state-of-the-art methods based on SD for the multiple conditional control tasks. In Section~\ref{subsec:ablation}, we further conducted ablation studies on the $\sgis$ and $\dcls$.

\subsection{Comparison with SD-based Methods}\label{subsec:compare}
\subsubsection{Skeleton Control Task}
we benchmark our model ECNet, against the most recent state-of-the-art (SOTA) model, HumanSD~\cite{ju2023humansd}, and the widely recognized SD-based control model, ControlNet~\cite{zhang2023adding}.  We use the LAION-Human dataset proposed in HumanSD as training set. The dataset is curated from the LAION~\cite{DBLP:journals/corr/abs-2111-02114} dataset, focusing on images of the highest quality and those that have received strong approval from human evaluators. It comprises 760,000 human image-text pairs. For validation, we choose HumanArt~\cite{ju2023human} dataset, which contains 4,750 images from natural and artificial scenes. This dataset, with its clean posture and textual annotations, offers a precise quantitative assessment of postures. We retrained HumanSD on the LAION-Human dataset for fairness, and the reported metrics are based on the Human-Art validation set.

We validate the performance of the ECNet in terms of posture generation quality and semantic association. We employ five metrics to evaluate the model: the distance-based Average Precision (AP), representing the similarity in human keypoint distances; the Pose Cosine Similarity-based AP (CAP), indicative of human posture similarity; the People Count Error (PCE), reflecting the accuracy of generated human figures; CLIPSIM, measuring the relevance between image information and textual descriptions; and the Fréchet Inception Distance (FID), which evaluates the quality of image generation.

\begin{figure}[t]
  \centering
   \includegraphics[width=1.0\linewidth]{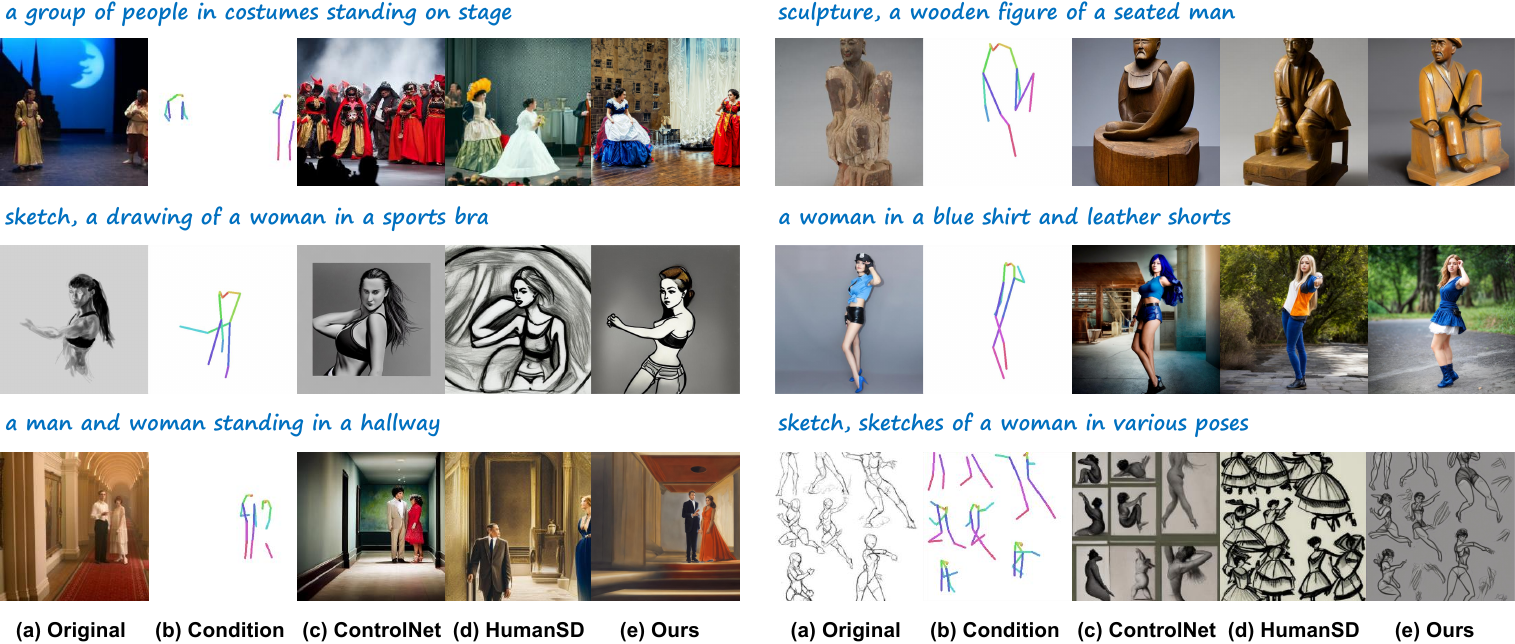}
   \caption{Generated images on the skeleton control task. The comparison of generated results across various scenarios validates the prior adaptability of ECNet.} 
   \label{pose sample}
\end{figure}

The results are shown in the left part of Table~\ref{tab:combined_metrics}. ECNet demonstrates the highest similarity in human keypoint distances with an AP score of 0.433, surpassing ControlNet and HumanSD, indicating its superior ability to capture and replicate the spatial configuration of human figures. With the highest CAP score of 0.628, ECNet has the ability of more accurate representation of human poses. ECNet has the lowest PCE value of 1.354 reflects its precision in generating multiple-human figures. ECNet scores CLIPSIM of 32.28, which, although slightly lower than HumanSD, still demonstrates a strong correlation between the generated images and their textual descriptions. With an FID score of 4.88, ECNet's image generation is close to the SOTA method HumanSD and ControlNet. More qualitative comparisons with HumanSD and ControlNet are illustrated in \Figref{pose sample}. The figure displays generated results across various scenarios, including sketches, sculptures, natural human, and multiple people, which represents the adaptability of ECNet across scenarios.

\subsubsection{Facial Landmarks Control Task}

\begin{figure}[t]
  \centering
   \includegraphics[width=1.0\linewidth]{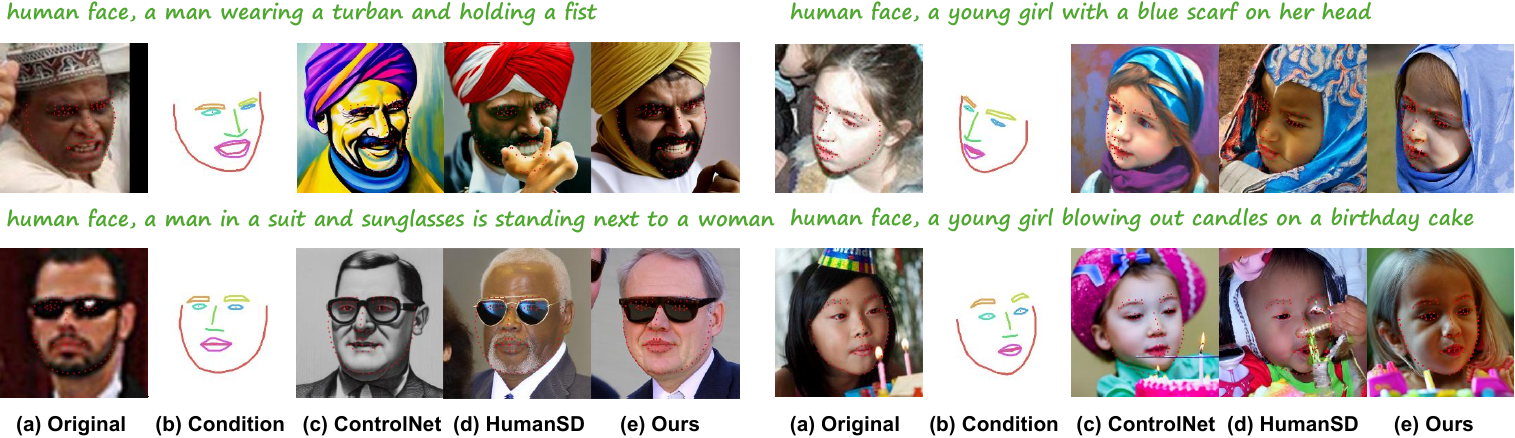}
   \caption{Generated images on facial landmarks control task. The comparison of generated results based on landmark control validates ECNet surpasses former SD-based models in this task.} 
   \label{facesample}
\end{figure}

In this section we compare ECNet with our baseline model HumanSD and ControlNet on facial landmarks control task. We utilize the high-quality FFHQ~\cite{DBLP:journals/corr/abs-1812-04948} facial dataset as our training set, comprising 70,000 high-definition facial images at a resolution of 1024x1024, showcasing diversity in age, ethnicity, and facial attributes. Then we construct the objective function through the landmark heatmaps of both the model-generated images and the input images, to apply supervision on the model training process. For validation, we employed a subset of 2,500 images from the WFLW~\cite{wayne2018lab} dataset, annotated with 98-point landmarks, to quantitatively assess the precision of generated faces. Note that the text pairs for the images are generated through a pretrained Bootstrapped Language Image Pretraining(BLIP) model~\cite{DBLP:journals/corr/abs-2201-12086}. As a benchmark, we also retrained HumanSD using the same datasets.
We evaluated the performance of ECNet in terms of accuracy of landmarks, semantic relevance, and image quality. The model's performance is assessed using three evaluation metrics: Normalized Mean Error (NME) based on distance for assessing the accuracy of generated faces, CLIPSIM for measuring the correlation between image information and text descriptions, and FID for evaluating the quality of image generation.
As illustrated in the right part of~\Cref{tab:combined_metrics}, the NME scores indicate that ECNet achieves significantly higher accuracy in the facial landmarks than the baseline, suggesting our training strategy is equally effective for controlling facial generation tasks.
Additionally, the reduction in FID scores highlights an improvement in the generation quality achieved by ECNet.
More qualitative comparisons are shown in~\Figref{facesample}, comparing with ControlNet and HumanSD, ECNet showcases superior performance in facial landmarks control tasks, balancing precise control with high quality generation. 


\subsubsection{Sketch Control Task}

\begin{figure}[t]
  \centering
   \includegraphics[width=1.0\linewidth]{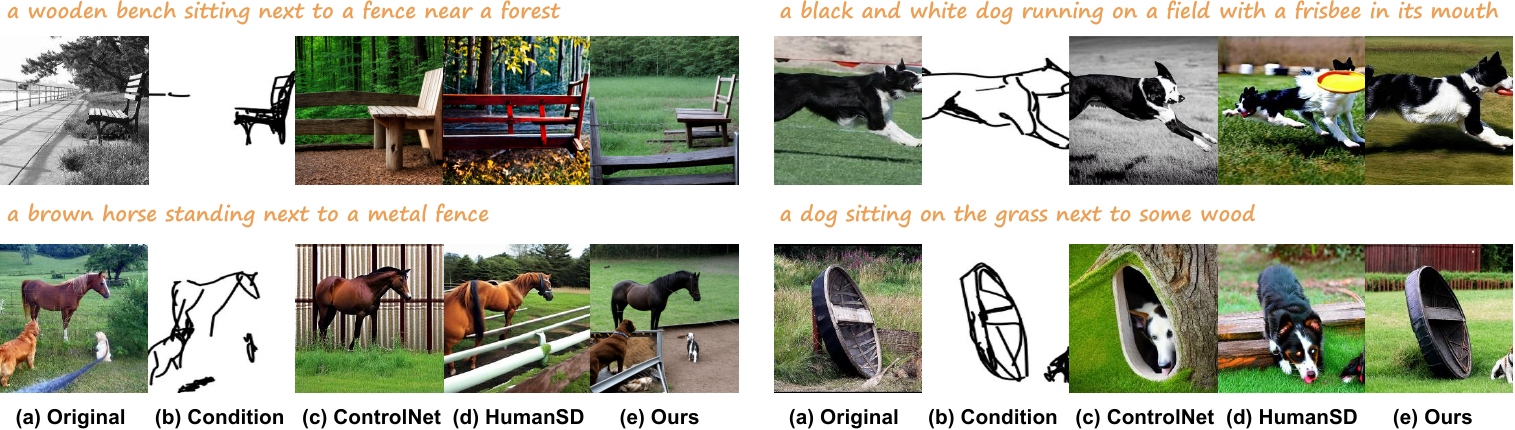}
   \caption{Generated images on the sketch control task. The comparison of generated results based on sketch control validates ECNet surpasses former SD-based models in this task.} 
   \label{sketchsample}
\end{figure}

In this section, we validate the effectiveness of the ECNet in the sketch control task. Initially, we employ CLIPasso~\cite{vinker2022clipasso}, a model capable of converting an image of an object into a sketch, to generate paired sketches for 5,000 image-text pairs across ten categories from the SketchyCOCO~\cite{gao2020sketchycoco} dataset, including airplane, bench, boat, cow, dog, elephant, horse, giraffe, train, and zebra. Subsequently, we extract 90 points from strokes per sketch as annotations to serve as input for the $\sgis$ module. thus far, we construct a small annotated sketch-paired image-text dataset, comprising 4,000 for training and 1,000 for validation. Finally, we sample 90 points from the attention distribution of the image generated by the model, aligned with the shape of the input annotations, used for $\dcls$ calculation in ECNet training. Qualitative comparisons with ControlNet and HumanSD, as shown in \Figref{sketchsample}, demonstrate that ECNet outperforms previous SD-based models in sketch-control generation.

\subsection{Ablation Study}\label{subsec:ablation}

In this section, we demonstrate the effectiveness of $\sgis$ and $\dcls$, which comprise $\mathcal{L}_{drv}$ and $\mathcal{L}_{dff}$, through the ablation study on the skeleton control task.

\subsubsection{Impact of Annotation Addition}

To verify the validation of the annotation addition module, we jointly train the $\sgis$ module with the original objective function of our baseline model HumanSD. This module, through the attention mechanism in the transformer structure, integrates spatial information such as the number of objects and keypoint locations contained in the annotation information (Anno.) into the textual conditions. This integration enables a more refined control over the generated results. As illustrated in the second row of Table~\ref{tab:ablation}, the integration of the $\sgis$ module enhances AP, CAP, and PCE scores, compared to the baseline model. This improvement underscores the effectiveness of incorporating such information in boosting the performance of human pose generation tasks. Furthermore, a reduction in the FID score suggests a slight improvement in the quality of images generated following the integration of the $\sgis$ module. Although incorporating annotation information into text conditions does have a minor adverse effect on text features, leading to a slight decrease in the CLIPSIM index, this impact is not substantial.

\begin{table}[t]
\centering
\setlength{\tabcolsep}{3pt}
\caption{Metrics for the ablation study, performances of baseline model, annotation addition, and guidance loss impact}
\begin{tabular}{lccccc}

\hline
\multicolumn{1}{l}{Model} & AP(\%)$\uparrow$ & CAP(\%)$\uparrow$ & PCE$\downarrow$ & CLIPSIM$\uparrow$ & FID$\downarrow$ \\
\hline
Base & 33.15 & 59.38 & 1.43 & \textbf{32.63} & 4.74\\
$\sgis$ & 37.90 & 60.27 & 1.37 & 32.08 & \textbf{4.58} \\
$\sgis \& \mathcal{L}_{drv}$ & 38.51 & 61.43 & 1.35 & 32.29 & 4.72 \\
Full & \textbf{43.31} & \textbf{62.76} & \textbf{1.35} & 32.28 & 4.89 \\
\hline
\end{tabular}
\label{tab:ablation}
\end{table}

\begin{figure}[t]
  \centering
   \includegraphics[width=\linewidth]{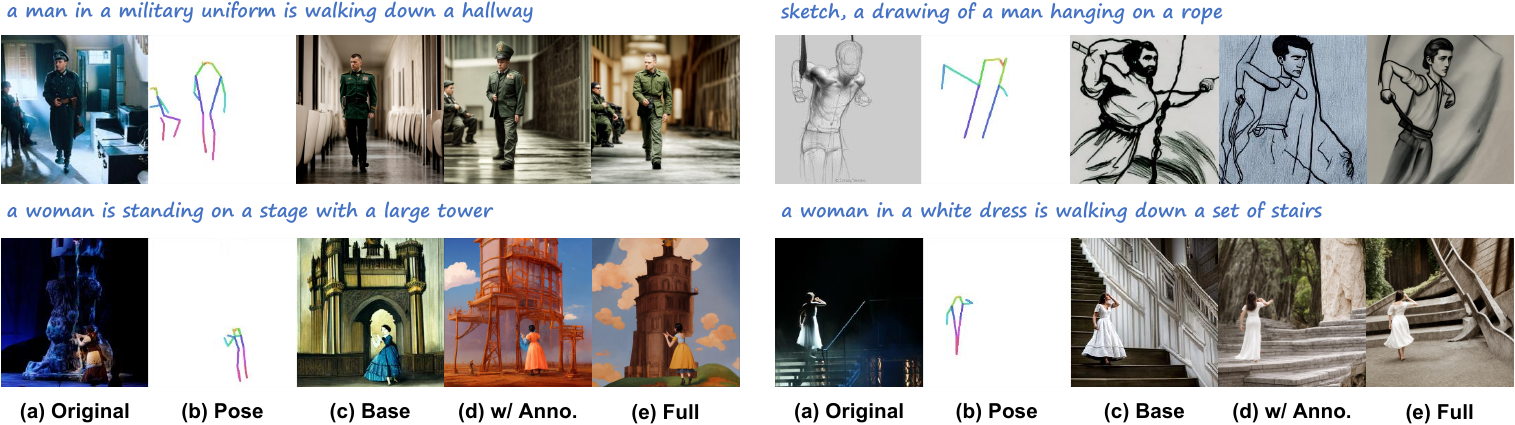}
   \caption{Generated results of baseline model,  with the annotation addition, and with the annotation plus loss supervision.}
   \label{ablation anno addition vis}
\end{figure}




\subsubsection{Impact of different losses}

In this section, we conduct validation experiments under identical conditions for the two losses proposed in the previous section: $\mathcal{L}_{drv}$ and $\mathcal{L}_{dff}$. As shown in the third row of Table \ref{tab:ablation}, the human pose metrics of $\mathcal{L}_{drv}$ surpass the model with $\sgis$ module. It is demonstrated that supervision on denoised latent code achieves more precise control in human pose generation. Following the incorporation of $\mathcal{L}_{dff}$, as shown in the fourth row of Table \ref{tab:ablation}, there is a further improvement in the metrics assessing human pose accuracy. This demonstrates that our proposed dual-stage loss, $\dcls$, can effectively mitigate the issue of larger errors at larger time steps encountered by $\mathcal{L}_{drv}$.

\Figref{ablation anno addition vis} illustrates the effect of applying $\sgis$ module and $\dcls$ in skeleton-based image generation. For each set of five images within the figure, from left to right, they represent: the original image, the pose condition, the generation by baseline model, the generation using the $\sgis$ module alone, and the generation by ECNet. The first pair of image sets displayed in the top row, illustrates ECNet's exceptional capability in handling scenarios involving multiple persons and in accurately generating images of rare poses. The second pair of image sets, displayed in the bottom row, highlights the efficacy of the $\sgis$ module in tackling the complex task of recognizing pose orientations and demonstrates how the $\dcls$ contributes to more precise pose control.

\section{Conclusion}

In this work, we introduce a novel framework ECNet that builds upon a pre-trained Stable Diffusion (SD) model. ECNet significantly enhances the generation of controllable models by incorporating $\dcls$ for consistency supervision on the denoised latent code of the diffusion model.Additionally, we enhance the model's perception of the ambiguities of input conditions by introducing Spatial Guidance Injector. This framework is designed to be versatile, preserving the generative capability of the pre-trained SD model while enhancing the influence of various input conditions on the outputs. In our comparative analysis with baseline models, employing diverse evaluation metrics such as posture and facial landmark precision, image quality, and relevance to text, ECNet demonstrably surpasses existing state-of-the-art models.


{
    \small
    \bibliographystyle{splncs04}
    \bibliography{egbib}
}


\end{document}


\title{ECNet: Effective Controllable Text-to-Image Diffusion Models}
\subtitle{-- Supplementary Material --}

\titlerunning{ECNet}

\author{}

\authorrunning{Li et al.}

\institute{}

\maketitle

\section{Additional Results}

In this section, we present more quantitative results for three tasks: skeleton control, facial landmark control, and sketch control.

\subsection{Skeleton}

\begin{figure}[th]
    \centering
    \begin{subfigure}[b]{0.9\textwidth}
        \includegraphics[width=\textwidth]{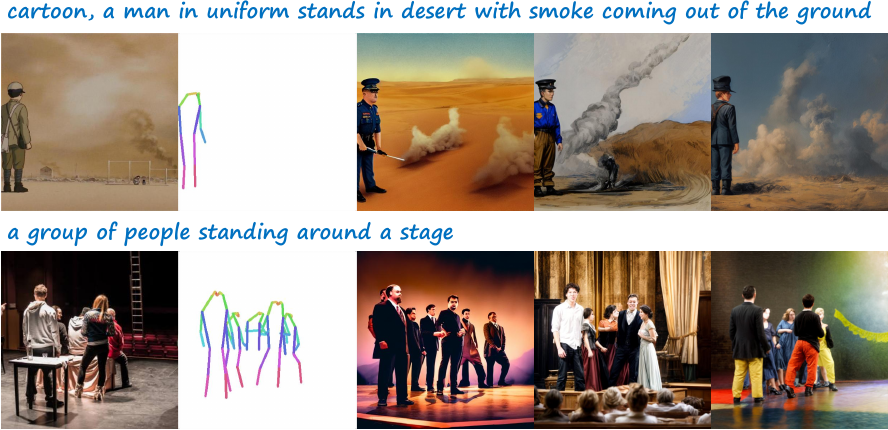}
        \label{pose_1}
    \end{subfigure}

    \vspace{-7pt}

    \begin{subfigure}[b]{0.9\textwidth}
        \includegraphics[width=\textwidth]{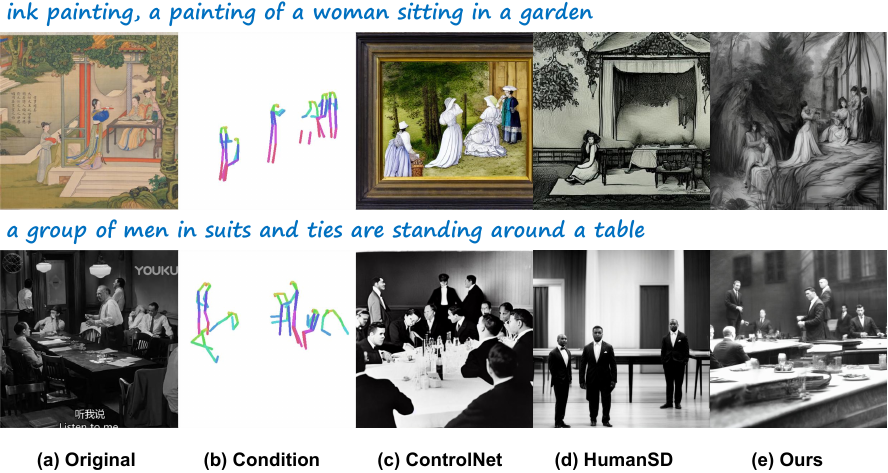}
        \label{pose_2}
    \end{subfigure}
    \caption{Comparison results with ControlNet and HumanSD in pose orientation recognition and multiple people scenario.}
    \label{fig:pose1}
\end{figure}

\begin{figure}[h]
    \centering
    \begin{subfigure}[b]{0.9\textwidth}
        \includegraphics[width=\textwidth]{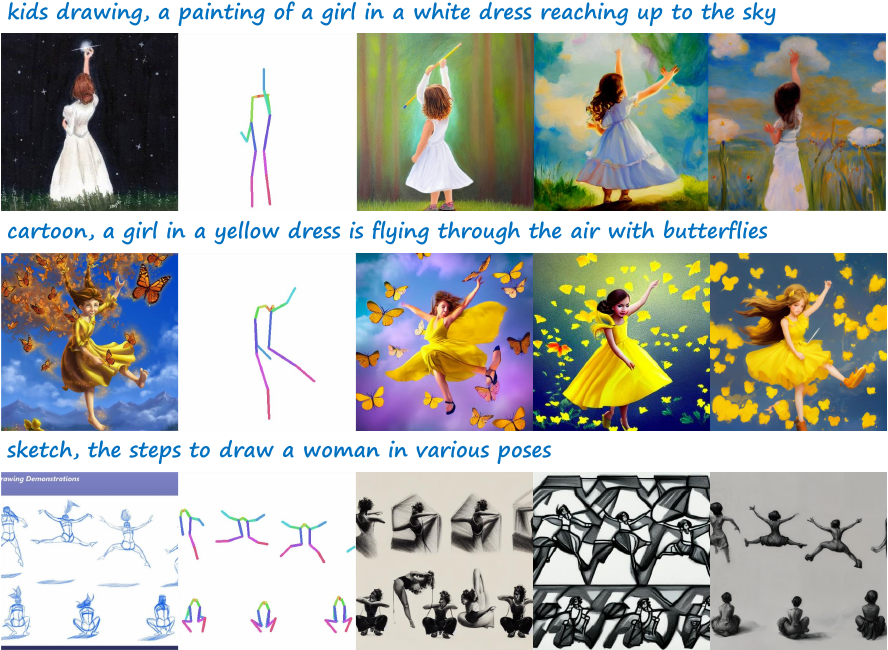}
        \label{pose_a}
    \end{subfigure}

    \vspace{-7pt}

    \begin{subfigure}[b]{0.9\textwidth}
        \includegraphics[width=\textwidth]{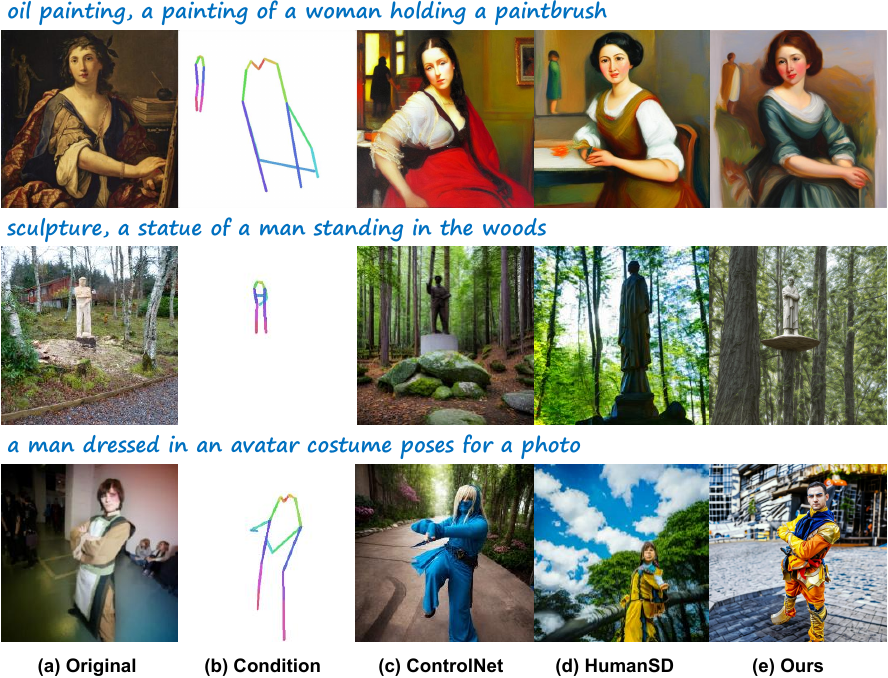}
        \label{pose_b}
    \end{subfigure}
    \caption{In the skeleton control task, more quantitative comparison results with ControlNet and HumanSD across various scenarios.}
    \label{fig:pose2}
\end{figure}

\begin{figure}[h]
    \centering
    \begin{subfigure}[b]{0.9\textwidth}
        \includegraphics[width=\textwidth]{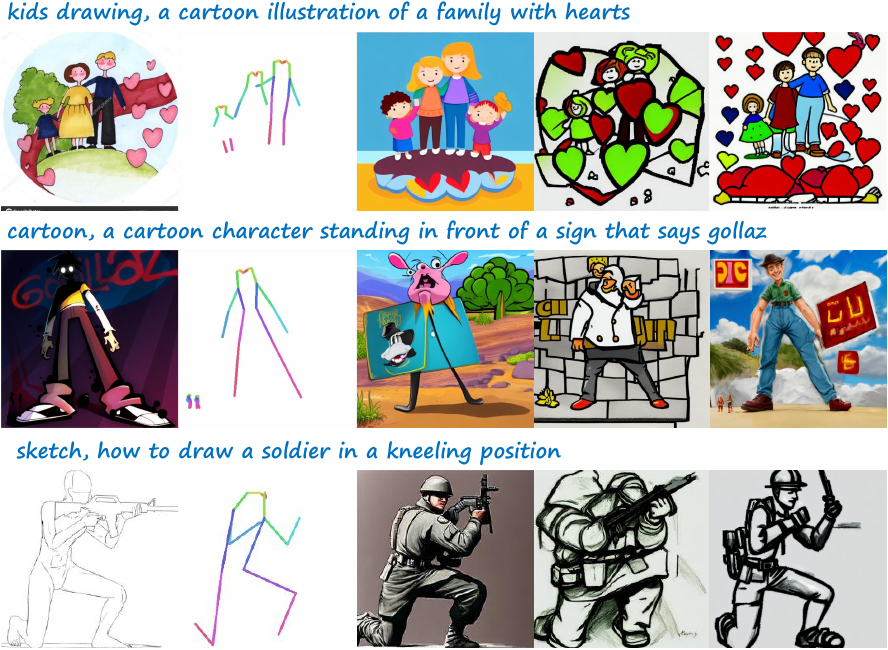}
        \label{pose_c}
    \end{subfigure}

    \vspace{-7pt}

    \begin{subfigure}[b]{0.9\textwidth}
        \includegraphics[width=\textwidth]{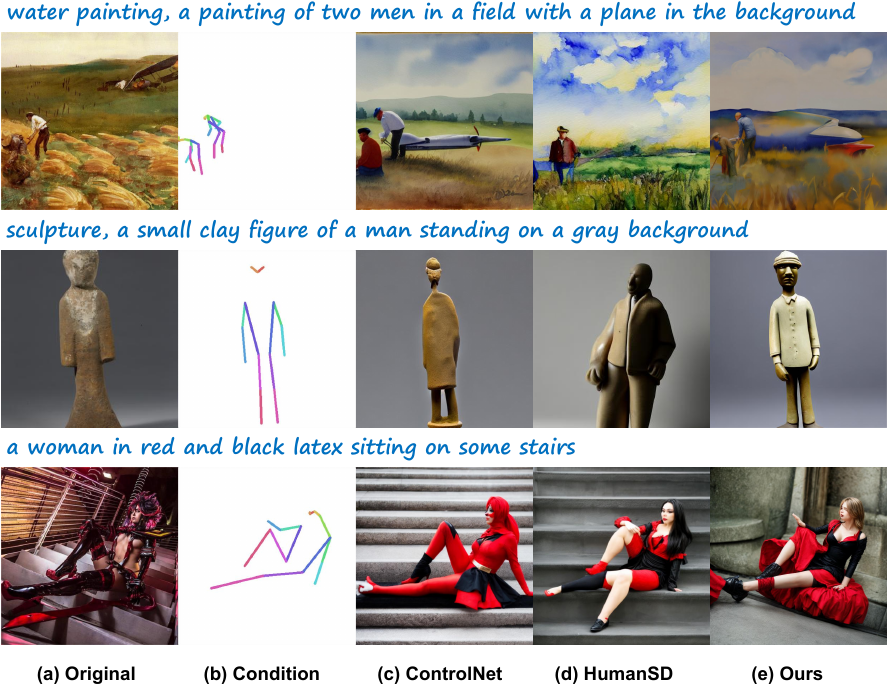}
        \label{pose_d}
    \end{subfigure}
    \caption{More quantitative comparison results with ControlNet and HumanSD in skeleton control task.}
    \label{fig:pose3}
\end{figure}

For the skeleton control task, we use the HumanArt dataset, including multi-scenario, human-centric images, as the validation set. Hence, we show comparison results of our model with ControlNet~\cite{zhang2023adding} and HumanSD~\cite{ju2023humansd} across various scenarios. This includes complex generation tasks like pose orientation recognition and multi-person scenes, as well as generation results in different styles such as cartoon, kid drawing, sketch, oil painting, sculpture, and natural human. The comparison results are illustrated in Fig.\ref{fig:pose1}, Fig.\ref{fig:pose2}, and Fig.\ref{fig:pose3}. From the results, our method demonstrates superior conformance to the input condition across various scenarios, yielding a high level of accuracy in pose replication. Notably, as illustrated in Fig.\ref{fig:pose1}, our model successfully discerns and reconstructs complex poses, such as back views, where existing methodologies fall short. Furthermore, in contexts involving multiple figures, our approach outperforms competitors by adhering closely to the condition, ensuring precise pose rendering and condition understanding. The enhanced performance of our model across different scenarios is attributable to the integration of SGI within the annotation input and the employment of a more robust and consistent DCL supervision mechanism. 






\subsection{Facial Landmark}

\begin{figure}[h]
  \centering
   \includegraphics[width=0.9\linewidth]{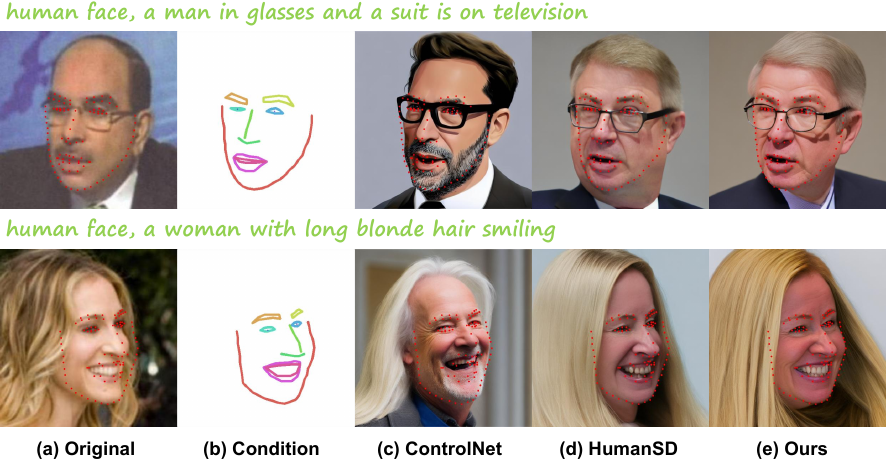}
   \caption{Some generated results according to inaccurate ground truth.} 
   \label{fig:face_c}
\end{figure}

\begin{figure}[h]
    \centering
    \begin{subfigure}[b]{0.9\textwidth}
        \includegraphics[width=\textwidth]{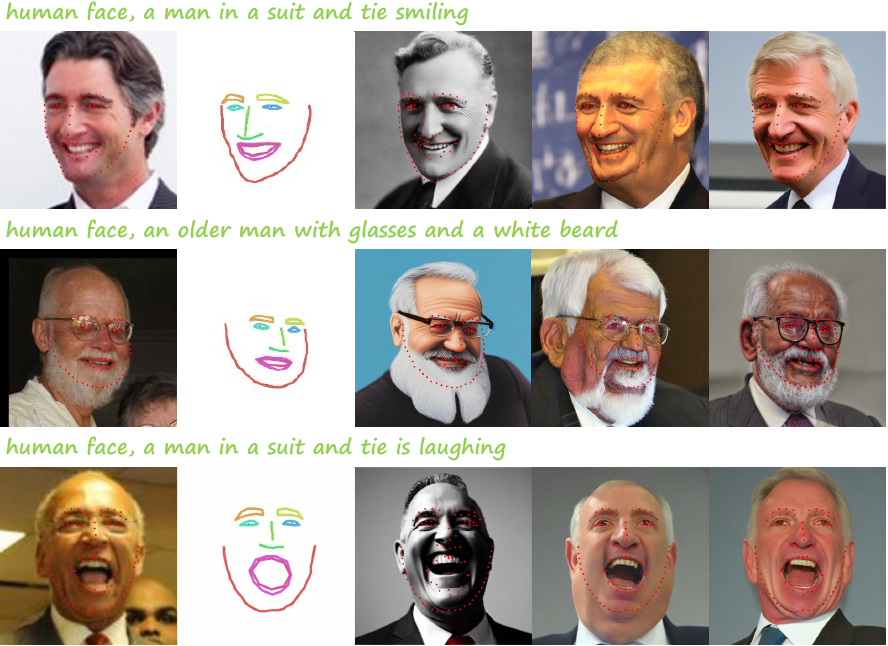}
        \label{face_a}
    \end{subfigure}

    \vspace{-5pt}
    
    \begin{subfigure}[b]{0.9\textwidth}
        \includegraphics[width=\textwidth]{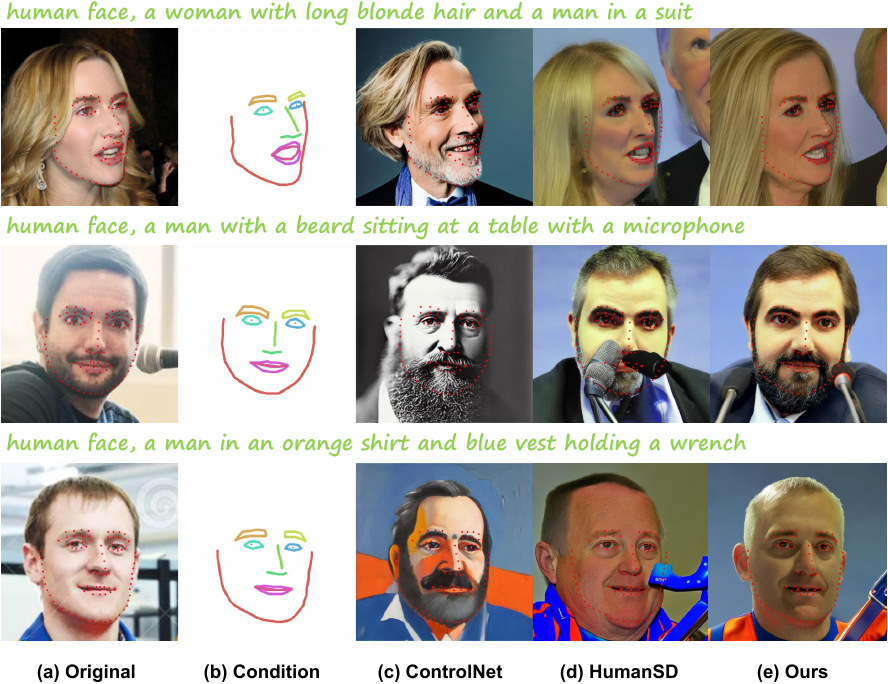}
        \label{face_b}
    \end{subfigure}
    \caption{More quantitative comparison results with ControlNet and HumanSD in facial landmark control task.}
    \label{fig:face_sup}
\end{figure}

 In the facial landmark control task, our quantitative comparisons, as depicted in Fig.\ref{fig:face_c} and Fig.\ref{fig:face_sup}, show that our model outperforms existing models significantly. We selected image pairs representing various facial expressions to assess the performance. Our approach demonstrates superior precision in aligning with the conditioned landmarks, underscoring the effectiveness of our end-to-end supervision strategy, which leverages SGI and DCL. Significantly, as illustrated in Fig.\ref{fig:face_c}, our model demonstrates the capability to generate images that adhere to the specified control conditions, even in instances where the original image does not align with these conditions. This precision in landmark adherence evidences the robustness of our method in capturing and replicating nuanced facial expressions.

\subsection{Sketch}

\begin{figure}[h]
    \centering
    \begin{subfigure}[b]{0.9\textwidth}
        \includegraphics[width=\textwidth]{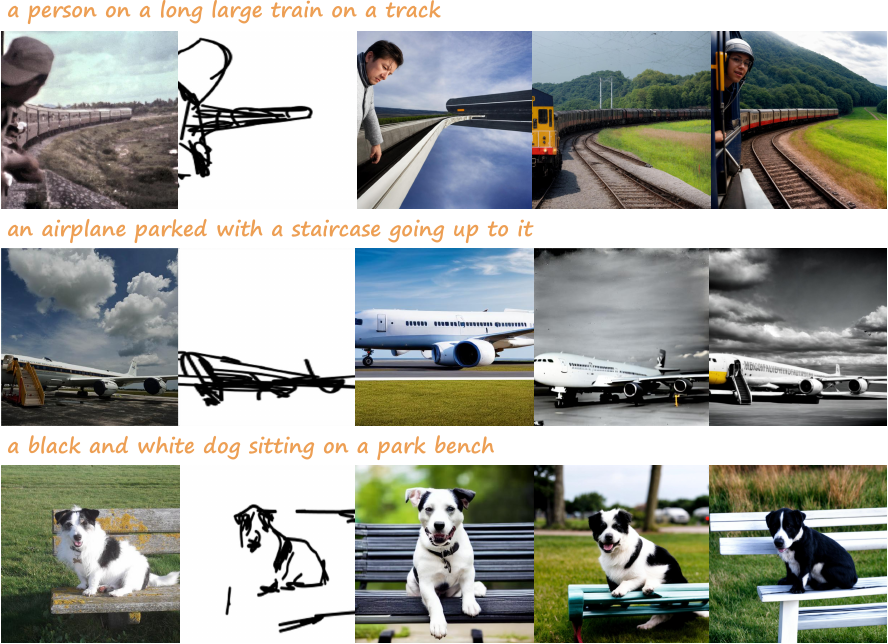}
        \label{sketch_a}
    \end{subfigure}

    \vspace{-7pt}

    \begin{subfigure}[b]{0.9\textwidth}
        \includegraphics[width=\textwidth]{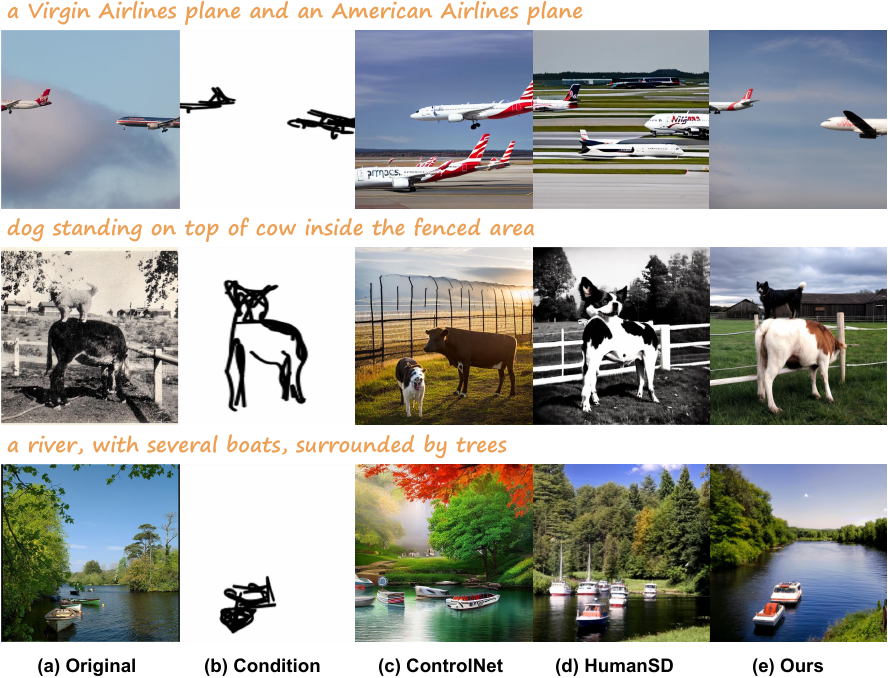}
        \label{sketch_b}
    \end{subfigure}
    \caption{More quantitative comparison results with ControlNet and HumanSD in sketch control task.}
    \label{fig:sketch_sup}
\end{figure}

For the sketch control task, we selected some cases involving man-made objects and animals, showing more quantitative comparison results. As shown in Fig.\ref{fig:sketch_sup}, these results prove our model outperforms other SD-based models in this task.

\section{Quality Enhancement}
Our training employs the pre-trained model stable-diffusion-v-1-5~\cite{Rombach_2022_CVPR} and the LAION~\cite{DBLP:journals/corr/abs-2111-02114} dataset for training. Due to the limitations in the generative capabilities of the pre-trained model and the image quality of the training set, achieving high-fidelity image generation is challenging. To enhance the quality of images in our paper, we utilize Tile ControlNet, a model within ControlNet designed for high-fidelity image restoration and quality improvement while preserving the overall structure of the images.

\section{Limitation and Future Work}
\begin{figure}[t]
  \centering
   \includegraphics[width=1\linewidth]{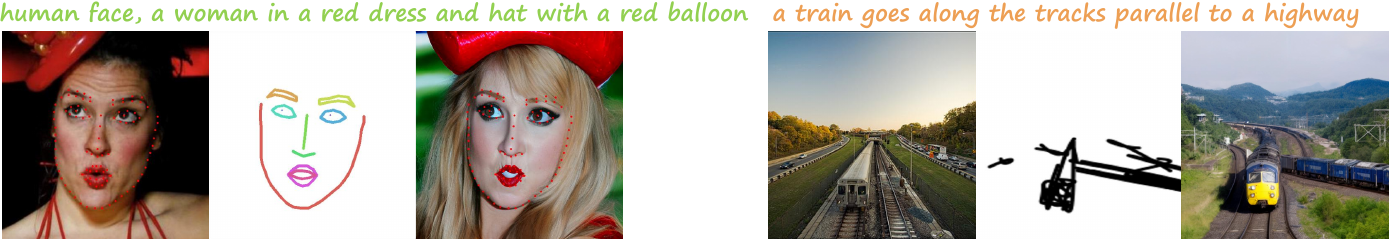}
   \caption{Some failure cases with lower semantic relevance. In the facial landmark case, the generated results do not include the "red balloon" mentioned in the prompt; in the sketch case, the generated results lack the "highway" semantic.} 
   \label{fig:failcase}
\end{figure}

Despite our proposed ECNet enhances controllability, it also faces certain limitations: (1) The model's supervision relies partly on detector performance, meaning annotation detection failures can impede supervisory capabilities. (2) The framework utilizes annotations as extra information added to the textual condition. This approach boosts control but lowers the relevance between image and prompt, some failure cases as illustrated in Fig.\ref{fig:failcase}. We plan to explore more robust methods of inserting annotations that balance both the model's control capabilities and semantic relevance. (3) The evaluation process remains limited, lacking verification under more conditions and scenarios. In the future, we plan to validate the effectiveness of our method on more tasks, such as segmentation layer control.

\bibliographystyle{splncs04}
\bibliography{egbib}
